\pdfoutput=1

\documentclass[11pt]{article}

\usepackage[final]{acl}

\usepackage{times}
\usepackage{latexsym}

\usepackage[T1]{fontenc}

\usepackage[utf8]{inputenc}
\usepackage{subcaption}

\usepackage{microtype}
\usepackage{hyperref}

\usepackage{inconsolata}

\usepackage{graphicx}
\usepackage{bbding}
\usepackage{pifont}
\usepackage{xcolor}
\usepackage{amssymb}
\usepackage{enumitem}
\usepackage{amsmath}

\title{Forgetting Curve: A Reliable Method for Evaluating Memorization Capability for Long-context Models}

\author{
  \textbf{Xinyu Liu\textsuperscript{1,*}},
  \textbf{Runsong Zhao\textsuperscript{1,*}},
  \textbf{Pengcheng Huang\textsuperscript{1}},
  \textbf{Chunyang Xiao},
  \\
  \textbf{Bei Li\textsuperscript{2}},
  \textbf{Jingang Wang\textsuperscript{2}},
  \textbf{Tong Xiao\textsuperscript{1,3}},
  \textbf{Jingbo Zhu\textsuperscript{1,3,$\dagger$}}
  \\
  \textsuperscript{1} \normalsize{NLP Lab, School of Computer Science and Engineering, Northeastern University, Shenyang, China} \\
  \textsuperscript{2} \normalsize{Meituan Inc.}
  \textsuperscript{3} \normalsize{NiuTrans Research, Shenyang, China} \\
  \normalsize{\{2310737, 2472148, 2201802\}@stu.neu.edu.cn, chunyangx@gmail.com} \\
  \normalsize{\{libei17, wangjingang02\}@meituan.com, \{xiaotong, zhujingbo\}@mail.neu.edu.com}
}

\begin{document}
\maketitle


\renewcommand{\thefootnote}{\fnsymbol{footnote}}
\footnotetext[1]{These authors contributed equally to this work.}
\renewcommand{\thefootnote}{\dag}
\footnotetext[2]{Corresponding author.}
\renewcommand{\thefootnote}{\arabic{footnote}}

\begin{abstract}
Numerous recent works target to extend effective context length for language models and various methods, tasks and benchmarks exist to measure model's effective memorization length. However, through thorough investigations, we find limitations for currently existing evaluations on model's memorization capability. We provide an extensive survey for limitations in this work and propose a new method called forgetting curve to measure 
the memorization capability of long-context models. We show that forgetting curve has the advantage of being robust to the tested corpus and the experimental settings, of not relying on prompts and can be applied to any model size.

We apply our forgetting curve to a large variety of models involving both transformer and RNN/SSM based architectures. Our measurement provides empirical evidence for the effectiveness of transformer extension techniques while raises questions for the effective length of RNN/SSM based models. We also examine the difference between our measurement and existing benchmarks as well as popular metrics for various models. Our code and results can be found at \href{https://github.com/1azybug/ForgettingCurve}{https://github.com/1azybug/ForgettingCurve}.

\end{abstract}
\section{Introduction}

\begin{figure}[hbt!]
    \centering
    \includegraphics[width=0.48\textwidth]{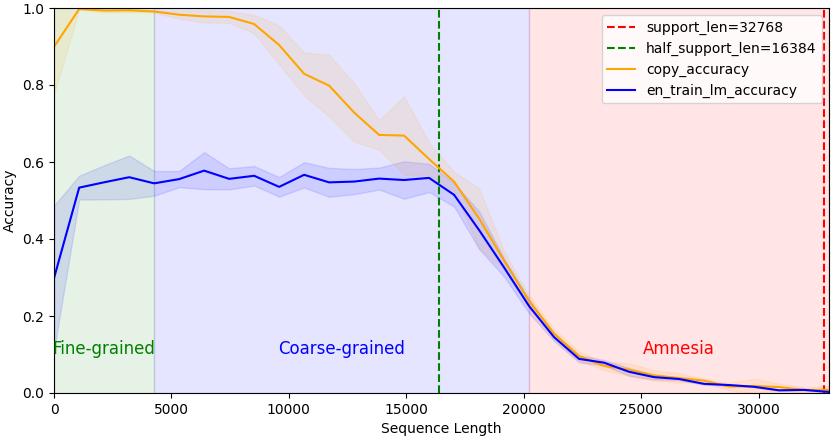} 
    \caption{The forgetting curve of Llama-2-base-32k~\cite{llama2_32k}. The x-axis denotes the prefix length. Green, blue, and red areas respectively indicate fine-grained memory where the model achieves 99\% token replication accuracy (except for very short sequences), coarse-grained memory where copy accuracy surpasses LM accuracy, and the amnesia area where the model completely ignores the prefix.}
    \label{fig:forgetting_curve}
\end{figure}

Arguably driven by large language model (LLM) practical needs, numerous works today aim to extend LLM contexts. The approaches range from modifying existing transformer architectures~\cite{dai2019transformer,wu2022memorizing,munkhdalai2024leave}, training existing LLMs to extend their contexts~\cite{chen2023longlora,xiong2023effective,chen2023extending}, to more recent modelling based on fundamental architectural changes using RNNs or hybrid architectures~\cite{gu2023mamba,peng2023rwkv,lieber2024jamba}, etc. All of the approaches claim to significantly extend the context window of corresponding baseline models with empirical validations. However, to authors' best knowledge, the community has not established (and thus has not agreed on) a standard metric that can reflect model's inherent memory with respect to the long context, in other words, model's effective memory length.

The most commonly used metric for evaluating a model's natural language generation ability is Perplexity (PPL). It is generally assumed that a superior long-range perplexity suggests a more effective utilization of long context. However, recent findings indicate that a lower perplexity for long context does not necessarily lead to improved performance in downstream long context applications~\cite{hu2024can}. This raises questions about the validity of using perplexity as an indicator of a model's long-context memorization capability.\footnote{The same paper suggests that perplexity effectively indicates model performance on short rather than long sequences.}
Alternative metrics have been proposed to evaluate a model's long-term memory, including ``Needle in a Haystack'' \cite{LLMTest_NeedleInAHaystack} and its derived metrics \cite{DBLP:journals/corr/abs-2403-11802}. However, these metrics also present their own limitations. For instance, ``Needle in a Haystack'' is highly sensitive to the prompt used,\footnote{As demonstrated in experiments by Claude~\cite{anthropic_claude_2_1} and Kimi~\cite{kimi}.} and such sensitivity is attributed more to the varying instruction-following capabilities across different models, rather than differences in their memory capabilities. Other metrics \cite{tay2020long,bulatov2022recurrent, bulatov2023scaling}, which involve toy tasks, face challenges in generalizing to real-world settings and are susceptible to severe overfitting. We survey a significant number of popular metrics and benchmarks which measure model long-term memory in this work.

To address these limitations and provide a reliable measure of a model's long-term memory, we introduce a new method called \textit{forgetting curve} in this paper. Our method also begins with a language model trained on a real-world corpus. We then exploit the model's emerging copy capability, assessing whether the trained model can replicate its prefix as part of the completion process. In order to account for instances where the token prediction relies solely on the model's natural language modelling (LM) capabilities instead of long-term memorization, we additionally plot the LM accuracy curve. Figure~\ref{fig:forgetting_curve} illustrates our method, demonstrating the progression of a memory pattern from fine-grained to coarse-grained memory, and ultimately to complete memory absence.

The paper is organized as the following. We start with a critical evaluation of the limitations inherent in the metrics and tasks currently employed to measure a model's long-term memorization capability (Section~\ref{sec:drawback}). Following this, we detail the construction of the forgetting curve, highlighting its dual function in visualizing memory and enabling statistical calculation of memory length. We also establish the reliability of our method in assessing a model's memorization capability (Section~\ref{sec:ForgettingCurve}). Using our forgetting curve, we comprehensively examine the long-range memorization capabilities of various current models and architectures (Section~\ref{sec:examination}). We demonstrate that the forgetting curves effectively illustrate a model's memorization capability through visualizations, revealing varying patterns across different architectures and models. We also validate certain context extension techniques, and raise questions about some architectural claims.
Finally in analysis (Section~\ref{sec:analysis}), we show that forgetting curves can be applied to models of different sizes under different settings. Compared to existing effective length measurements~\cite{hsieh2024ruler} or perplexity, the forgetting curve provides a self-contained, memory-focused measurement.



\section{Limitations of Long-Range Memory Measures}
\label{sec:2}
\label{sec:drawback}

\paragraph{Limited Memory Usage (LMU)} The first limitation that we identify in the long-range measurement literature is limited memory usage (LMU), where the measurement or metric does not have a strong relationship with model's capability to leverage long context. Perplexity is a surprising example of LMU. Recent works~\cite{chen2023longlora,yang2023longqlora,xiao2023efficient} typically demonstrate their model's long-range capability by showing a stably low perplexity given long contexts. However, ~\citet{hu2024can} recently found that lower perplexity does not imply an improved downstream performance on long-context tasks. In section \ref{sec:analysis}, we further demonstrate, using forgetting curves, that some models can improve long-range perplexity without enhancing long-range memory. LMU also manifests in popular long-range measurement tasks such as ``Needle in a Haystack''\cite{LLMTest_NeedleInAHaystack} which assesses the capability to retrieve a specific piece of information from lengthy distracting texts. The memory usage is clearly limited since the model can theoretically achieve great performance by focusing only on the `needle' and ignoring all the long context. Methods derived from ``Needle in a Haystack'' like Counting Stars~\cite{DBLP:journals/corr/abs-2403-11802} alleviate this drawback but still inherit such limitations.

\paragraph{Prompt Required (PR)} The second limitation in existing evaluation methods is their strong dependency on the exact prompts during evaluation. For instance, in the Claude 2.1 experiments \cite{anthropic_claude_2_1}, a single prompt change dramatically improved the model's ability to find the `needle', boosting Claude 2.1’s task success rate from $27\%$ to $98\%$ in the original evaluation. Such significant dependency on prompts raises the following two issues. Firstly, it makes the comparisons between models extremely difficult due to the large performance variance across different prompts.\footnote{Treating all prompts as i.i.d experiments allows for theoretical comparison of two models using statistical tests. However, large variance makes achieving statistical significance challenging. Furthermore, the i.i.d assumption is questionable.} Secondly, it is challenging to apply these metrics to unaligned models and smaller ones due to their limited capability to follow instructions to find the `needle'. This implies that new architectures need to be scaled up and aligned before these metrics can be applied, preventing rapid and low-cost experimental validation of a model’s long-context capability. Although latest benchmarks such as Ruler~\cite{hsieh2024ruler} and StreamEval~\cite{xiao2023efficient} can force a model to produce answers through carefully designed prompts, this approach is still cumbersome and model-specific.

\paragraph{Inference Factors (IF)} The third limitation we identify is that current long-context evaluations often 
fail to distinguish the influence of inherent inference factors, such as natural language understanding ability, etc.
For instance, in the ``Needle in a Haystack'' task where the model is prompted to locate the needle, a failure can be difficult to diagnose, as it could be due to a deficiency in either the model's long-context memorization capability or its natural language understanding capability. Similarly, failure in tasks such as long context question answering or long mathematical expression calculation may not reflect shortcomings in the model's long-term memory, but rather, deficiencies in its specific abilities.

\paragraph{Limited Context (LC)} The fourth limitation arises from the limited context inherent in benchmarks. Some benchmarks ~\cite{li2024long,zhang2024infty,li2023loogle} conduct their tests within a confined context, which restricts their ability to evaluate model performance in longer contexts. Expanding these benchmarks to include extended contexts is typically laborious and time-intensive, posing a challenge to keep pace with the rapid increase in context length that large language models can handle.

\paragraph{Training on Toy Tasks (ToTT)} Finally, we notice that many recent studies demonstrate a model's long-range capability by training on toy tasks~\cite{tay2020long,gu2023mamba,bulatov2023scaling}. While we recognize its contributions, we also argue that because long-context memory is a capability that matters in real applications such as long document summarization~\cite{fan-etal-2019-eli5}, role playing~\cite{li2023chatharuhi,wang2023rolellm} etc., the tests should be performed on real data. In other words, we argue that long-context capability should be an emerging capability to be measured, which does not involve any specific training, especially not any training on toy datasets. Ideally, the measurement should also be on a real dataset to mitigate issues like overfitting and data leakage~\cite{wu2023condefects,balloccu2024leak}. 

\begin{table}[!]
\fontsize{8}{9.6}\selectfont
\begin{tabular}{ll}
\hline
     Metrics/Tests/Tasks/Benchmarks & Limitations \\
     \hline
     Perplexity & LMU \\
     \hline
     Needle In A Haystack~\cite{LLMTest_NeedleInAHaystack} & LMU, PR, IF \\
     Counting Starts~\cite{DBLP:journals/corr/abs-2403-11802} & LMU, PR, IF \\
     StreamEval~\cite{xiao2023efficient} & LMU, PR \\
     \hline
     Passkeys~\cite{munkhdalai2024leave} & ToTT, LMU, PR, IF \\
     Selective Copy~\cite{gu2023mamba} & ToTT, LMU \\
     Copy~\cite{bulatov2022recurrent} & ToTT \\
     LRA~\cite{tay2020long} & ToTT, LC \\     
     \hline
     BABILong~\cite{kuratov2024search} & LMU, PR, IF \\
     LongICLBench~\cite{li2024long} & PR, IF \\
     Ruler~\cite{hsieh2024ruler} & LMU, PR, IF \\
    LongBench~\cite{bai2023longbench}  & LC, PR, IF \\
    InfiniteBench~\cite{zhang2024infty} & LC, LMU, PR, IF \\
    LooGLE~\cite{li2023loogle}      & LC, PR, IF \\
    ZeroSCROLLS~\cite{shaham2023zeroscrolls} & LC, PR, IF \\
    L-Eval~\cite{an2023eval}      & LC, PR, IF \\
    BAMBOO~\cite{dong2023bamboo}      & LC, PR, IF \\
    LV-Eval~\cite{yuan2024lv}     & LC, LMU, PR, IF \\
    CLongEval~\cite{qiu2024clongeval}  & LC, LMU, PR, IF \\
\hline
\end{tabular}
\caption{Overview of long-context measurements and limitations according to the categorization in Section \ref{sec:drawback}.}
\label{tab_metrics_limitations}
\end{table}

\vspace{0.3cm}

\noindent Table~\ref{tab_metrics_limitations} lists popular methods for assessing a model's long-context performance along with their limitations. It is important to note that these limitations are often interrelated. Ideally, assessment methods should overcome all the listed limitations to provide an accurate measure of a model's performance.

\begin{figure*}[t]
\centering
\includegraphics[width=1.0\textwidth]{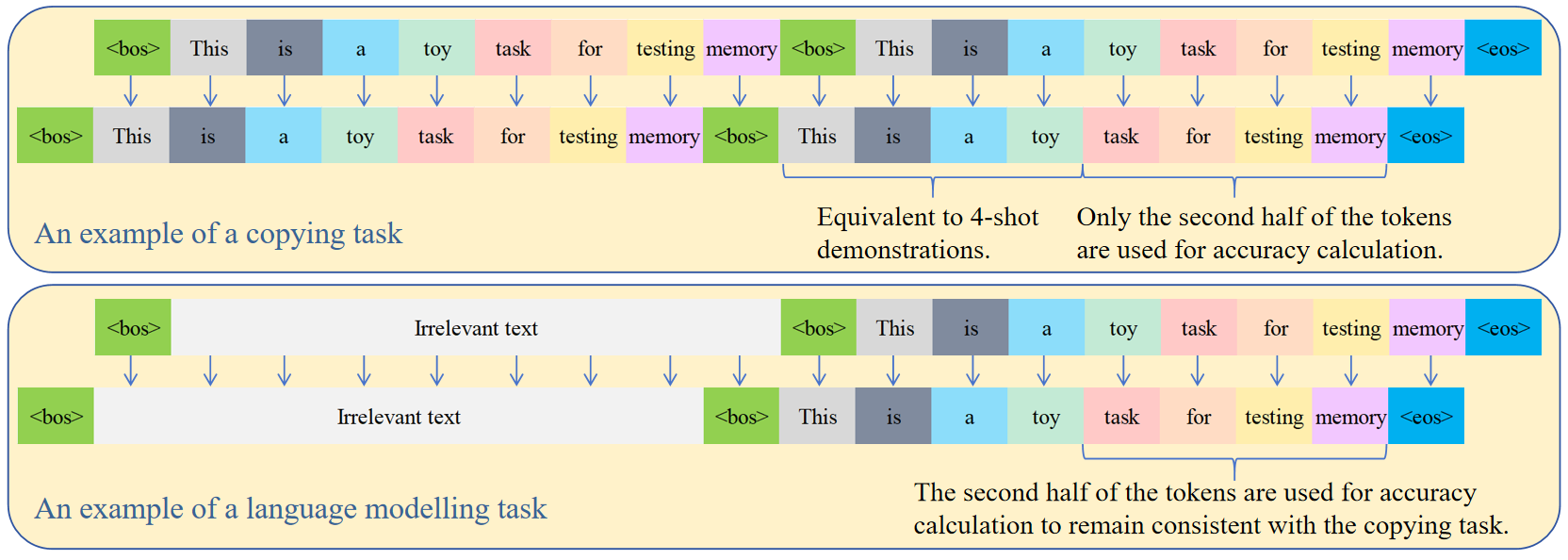}
\caption{The forgetting curve task measures the LLM prediction accuracy for the target sequence ``This is a toy task for testing memory'' under two settings. The above figure illustrates the copy setting, while the below one shows the language modelling setting. We calculate the difference between these two settings to obtain the forgetting curve reflecting model's memory behaviour. As shown in the figure, only the later half of the tokens are taken into account to construct the forgetting curve.}
\label{fig:forgetting_curve_method}
\end{figure*}

\section{The Forgetting Curve}
\label{sec:ForgettingCurve}


In this work, we introduce a novel method called ``forgetting curve'' to measure a model’s inherent long-context memorization capability. We show that the forgetting curve effectively mitigates the previously listed limitations and remains robust under various experimental settings. This makes it an ideal method for characterizing and visualizing a model's long-context memorization capability.

\subsection{Evaluation Methodology}
\label{subsec:evaluation_methodology}
The forgetting curve is derived from LLM copy tasks and consists of two curves (i.e copy accuracy curve and LM accuracy curve) as shown in Figure~\ref{fig:forgetting_curve}. For both curves, teacher forcing is applied to measure the next token prediction accuracy. We refer to teacher forcing the setting widely used to measure language model perplexity where the next token prediction probability is taken into account by always assuming a correct prefix derived from a corpus. We only differ from perplexity measurement in that we measure token prediction accuracy (i.e. 0 or 1 for each token) instead of log probability scores for both copy accuracy curve and LM accuracy curve. We further detail the measurement of the two curves below.   
\subsection{Evaluation Tasks}

\paragraph{Copy Accuracy Curve.} 
The first curve, depicted as the yellow curve in Figure~\ref{fig:forgetting_curve}, is the copy curve. This curve assesses an LLM based on its copy accuracy. We calculate this accuracy by tasking the model to predict the second half of a copied string using its first half, as illustrated at the top of Figure~\ref{fig:forgetting_curve_method}. One key difference from the standard copy task is our requirement for the model to be \textit{not trained} on such tasks, thus treating copy as an emergent capability of the LLM. Such constraints avoid trivial solutions which overfit to the copy tasks, as pointed out by~\citet{gu2023mamba}. Secondly, instead of using natural language prompts to trigger the LLM's copy behavior, we employ teacher forcing and measure the resulting predicted copy accuracy. The initial tokens are disregarded from the measurement as they are mainly used as few-shot learning examples to inform the model of the copy task.~\footnote{Remark that such way of indicating copy task (by showing the model the copied tokens) effectively avoids relying on model's understanding capability, making our method applicable for unaligned/smaller models as we show in Section~\ref{sec:analysis}.} In our experiments, all curve measurements begin with the copy performance for the second half of the target sequence. 
\paragraph{LM Accuracy Curve.} In the forgetting curve, we choose real-life natural language texts to avoid trivial or overfitting solutions to the copy task. However, this choice also implies a non-negligible probability of correctly predicting the token even without access to any long-range contexts.\footnote{One might notice that this is the same issue we highlighted in Section~\ref{sec:drawback} regarding perplexity.} To take this probability into account, the forgetting curve also plots the language modelling accuracy, represented by the blue curve in Figure~\ref{fig:forgetting_curve}. Specifically, an irrelevant but real text of the same length as the copy target is chosen.\footnote{In Section~\ref{sec:3.3.1}, we demonstrate that the distribution used to sample the random string does not impact the language model accuracy.} We measure the language model's accuracy in predicting the later half of the target sequence, using the irrelevant prefix and the first half of the sequence, as depicted in Figure~\ref{fig:forgetting_curve_method}.


\subsection{An Illustrative Example}
We run Llama-2-base-32k model~\cite{llama2_32k} using test text from PG-19~\cite{rae2019compressive} to produce a forgetting curve (Figure~\ref{fig:forgetting_curve}). In these experiments, we compute an accuracy score every 1K tokens for both the copy curve and the language model  curve. Each accuracy is calculated 10 times, allowing us to plot the mean accuracy and the variance in the curve.

We visualize the corresponding forgetting curve in
Figure~\ref{fig:forgetting_curve}. From the first accuracy data point (1K), we observe that Llama-2-base-32k performs the copy task perfectly without explicit instruction, a phase lasting until 4K.\footnote{We remark that the copy accuracy is not 100\% for the first measurement. The phenomenon is universal (see Appendix~\ref{appendix:forgetting_curves} for the forgetting curves of other models we measure) and we believe it is because at this stage, with relatively few examples demonstrating the copy behaviour, the model does not learn to copy perfectly yet.} We refer to this phase as \textit{fine-grained memory}, as the model can perfectly memorize past tokens.
After this phase, the model can no longer precisely remember all the previous context, as shown by the copy accuracy curve. However, the model still exhibits significantly higher copy accuracy.
We refer to this phase as \textit{coarse-grained memory}.\footnote{In our setting, fine-grained memory length is where the model can accurately copy more than 99\% tokens, while coarse-grained memory length is where copy accuracy exceeds language model accuracy by more than 1\%.} For Llama-2-base-32k, we notice that during the coarse-grained memory phase, the copy performance gradually degrades to the level of language modeling (LM) accuracy.\footnote{Note that in ideal case, one would expect coarse-grained memory to last much longer than model's claimed context length.}
Finally, we observe no difference in the model's performance based on whether the target sequence has been previously seen by the model. We refer to this as the model entering the \textit{amnesia} phase.
\subsection{In-depth Analyses on Forgetting Curves}
In this subsection, we critically assess the reliability of the forgetting curve as an indicator of a model's memorization capacity. 
Our evaluation focuses on two primary aspects: \ding{182} the effect of using various irrelevant prefixes when measuring language modeling accuracy, and  \ding{183} the impact of testing on different text corpora. Moreover, we investigate whether the forgetting curve mitigates the limitations highlighted in Section~\ref{sec:drawback}.

\begin{table}[!]
\begin{tabular}{lll}
\hline
     Setting & Text source & Distribution \\
     \hline
     en\_train & PG19-train & ID(En) \\
     en\_test & PG19-test & ID(En) \\
     zh & LooksJuicy/ruozhiba & ID(Zh) \\
     random & Random & OOD \\
\hline

\end{tabular}
\caption{We extract both irrelevant and target copying text from various sources. All texts from PG-19 and Ruozhiba, which includes English and generated Chinese responses by GPT-4 respectively, are natural language and hence considered in-distribution. In contrast, a random token sequence, not adhering to any natural language distribution, is deemed out-of-distribution.}
\label{tab_metrics_settings}
\end{table}

\paragraph{LM Accuracy with Various Irrelevant Prefixes}
\label{sec:3.3.1}
To examine the effect of various irrelevant texts on the language model accuracy, we experiment with varying irrelevant text prefixes, ranging from in-distribution (ID) to out-of-distribution (OOD) relative to the Llama-2-base model. We source these irrelevant texts from four distinct text corpora, as detailed in Table~\ref{tab_metrics_settings}, with the PG-19 test texts serving as the copy target in all experiments.

\begin{figure}
    \centering
    \begin{subfigure}{.48\textwidth}
        \centering
        \includegraphics[width=\textwidth]{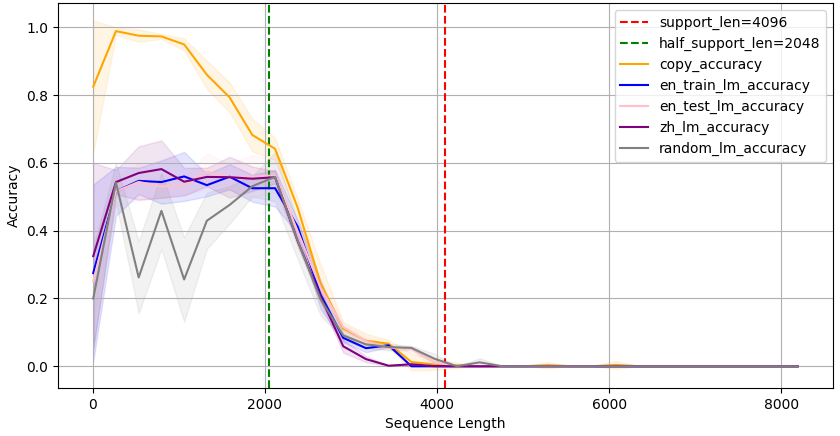} 
        \caption{} 
        \label{fig:diff_irrelevant}
    \end{subfigure}\hfill
    \begin{subfigure}{.48\textwidth}
        \centering
        \includegraphics[width=\textwidth]{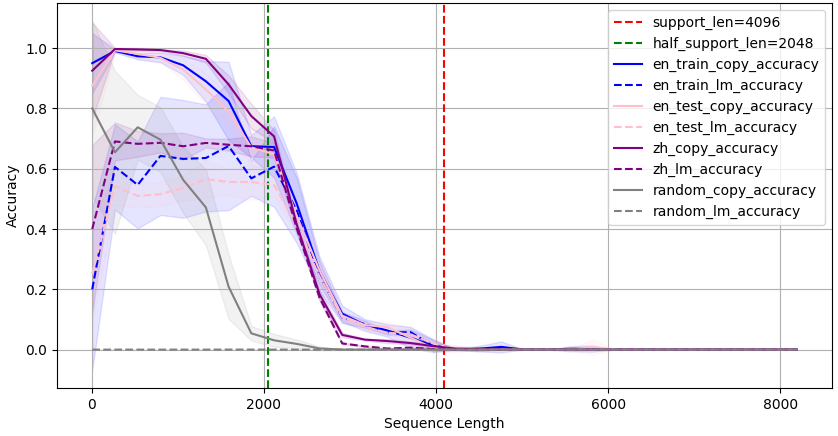} 
        \caption{} 
        \label{fig:diff_copy}
    \end{subfigure}
    \caption{Llama-2-7b forgetting curve with various text sources. (a) Various irrelevant text sources, and copy text is sourced from pg19 test set. (b) Various copy text sources, and irrelevant text is sourced from pg19 test set.}
    \label{fig:diff}
\end{figure}

The results are illustrated in Figure~\ref{fig:diff_irrelevant}. Except for the random prefix setting, all other settings yield similar language modeling accuracy, showing a significant downturn around half the claimed context length. These trends remain statistically indistinguishable up to the memorization threshold, as confirmed by our statistical analysis in Appendix~\ref{appendix:statistical_analysis}. Importantly, these observations suggest that the model's coarse-grained memory length is consistent across different prefix settings, demonstrating the robustness of the forgetting curve as a measure of memorization.




\paragraph{Forgetting Curves with Various Copy Targets}
We then assess the impact of copy text on the forgetting curve, using different copy targets sourced according to Table~\ref{tab_metrics_settings}.
The forgetting curves, depicted in Figure~\ref{fig:diff_copy}, show uniform trends across various settings, except when random targets are used. This uniformity in forgetting curves underscores a consistent coarse-grained memory length, highlighting the measure's reliability across different textual contexts. 

Our results confirm that the forgetting curve is a reliable measure of memorization, even when we change the irrelevant text at the beginning or the texts used in the copying task. This consistency holds as long as the texts come from well-organized sources and not just random samples. Essentially, the forgetting curve proves to be more dependable than expected, consistently showing similar patterns and lengths of memorization across different tests.

\paragraph{Analysis of Reliability} Our approach effectively addresses the limitations aforementioned in Section \ref{sec:2}. By requiring the language model to replicate all seen content in the copy task, we overcome the limitation of \textbf{LMU}. The inherent copy capabilities of language models, leveraged in our methodology, allow for the completion of the copy task using the copy text itself as a contextual cue, thereby circumventing the need for task-specific training \textbf{ToTT} or specialized prompts \textbf{PR}. Furthermore, the ability to construct a forgetting curve of arbitrary length eliminates the \textbf{LC} limitation. Finally, by comparing the language modeling accuracy and excluding prompt words from our evaluations, we minimizes the impact of inherent inference factors \textbf{IF}.

\section{Experiments}
\label{sec:examination}
\paragraph{Experimental Settings} We experiment with in total 14 open-source long-context LLMs and trace their forgetting curves to examine their memorization capability; the detailed information about the experimented models is included in Appendix~\ref{appendix:models}. The forgetting curve does not require natural language understanding capability and can be applied to any language models, allowing us to cover diverse model type (aligned or not aligned), model architecture (transformers or RNNs), and claimed context lengths (32k to 1M). For all models, we pull the model from their huggingface releases and run the inference in BFloat16 on 8 NVIDIA A100 GPUs. Both the copy accuracy curve and the LM accuracy curve are based on token prediction accuracy which is derived from the teacher forcing setting explained in subsection~\ref{subsec:evaluation_methodology}.


We utilize the PG19 test set~\cite{rae2019compressive} containing a collection of 100 books. Each book is individually tokenized based on the tokenizer settings of the model under test. To measure context of arbitrary length, we concatenate all books into an extremely long sequence. Subsequently, we divide the maximum supported length \textit{l}, as claimed by the model, into $n$ test lengths where $n$ is a fixed granularity (i.e we test model on length $(\dfrac{l}{n}, \dfrac{2l}{n}, ..., l)$ respectively).\footnote{We take n=32 for most of the cases to be able to trace a representative curve.} For each test length, we randomly select a continuous sequence from the concatenated text as a copy target while the remaining part is used to sample the irrelevant text to measure LM accuracy curve.
As shown in Figure~\ref{fig:forgetting_curve_method}, the copy task is presented to LM model in the form of $[bos] S [bos] S [eos]$ where $S$ is the target sequence to copy. For the language modeling task, LM accuracy is computed based on $[bos] I [bos] S [eos]$ where $I$ represents the irrelevant text. For each accuracy datapoint of certain context length, we sample 10 different copy target so that the curve contains mean and variance. 

\paragraph{Remarks on Closed Source Models} The forgetting curve can be applied to closed source models to measure their long context memorization capability. However, as the parameters of commercial models are not accessible, simulating the teacher forcing process requires feeding the prefix to the model and calculating the prediction accuracy for each token, which is too costly and inefficient to perform.

\begin{table*}[htp]
  \centering
  \begin{minipage}{\textwidth}
    \centering
    \begin{tabular}{l r r r r}
      \hline
      \textbf{Models} & \textbf{Claimed Length} & \textbf{Fine Length} & \textbf{Coarse Length} & \textbf{Effective Length} \\
      \hline
	Llama-2-base (7B) & 4k & 0 & 2.7k & criterion \\
	Llama-2-chat (7B) & 4k & 1.6k & 2.7k & criterion \\
	Llama-3-base (8B) & 8k & 4k & $>$9k & - \\
	Llama-3-chat (8B) & 8k & 4k & 6k & - \\
       \hline
 	Llama-2-base-32k (7B) & 32k & 4k & 20k & 4k \\
	Mistral-base-v0.2 (7B) & 32k & 22k & $>$30k & 16k \\
	Mistral-chat-v0.2 (7B) & 32k & 15k & $>$33k & 16k \\
 	Mistral-chat-v0.3 (7B) & 32k & 22k & $>$33k & - \\
	Qwen1.5 (7B) & 32k & 28k & $>$30k & - \\
	ChatGLM3 (6B) & 128k & 37k & $>$62k & 4k \\
	Yarn-Llama-2 (7B) & 128k & 0 & 82k & $<$4k \\
	LWM (7B) & 1M & $>$225k & $>$225k & $<$4k \\
      \hline  
	Mamba (2.8B) & 2k/infinite & 0 & 2.3k & $<$1k \\
	RWKV (7B) & 4k/infinite & 265 & 3.6k & 1k \\

      \hline
    \end{tabular}
    \caption{Performance of open-source models in Forgetting Curve and Ruler~\cite{hsieh2024ruler}. The Fine-grained Memory Length (Fine Length) is the maximum length at which copy accuracy surpasses 99\%. The Coarse-grained Memory Length (Coarse Length) is the maximum length at which the copy accuracy is greater than the language modelling accuracy by at least 1\%. In Ruler, the Effective Context Length (Effective Length) is the maximum length where the performance reaches a certain threshold, which is defined by the Llama-2-7b-base/chat performance at 4K. In the Fine length/Coarse Length columns, if there is a ``$>$'' symbol, it signifies that the length exceeds the indicated measure.}
    \label{tab:main_results}
  \end{minipage}
\end{table*}

\paragraph{Results} Table~\ref{tab:main_results} shows our main results, listing the key statistics derived from forgetting curves of each model. The exact forgetting curves can be found in figures from Appendix~\ref{appendix:forgetting_curves}. The models are grouped into three families: Llama models without context extensions, Transformer based models with context extensions and SSM/RNN models. 

We observe a clear memorization capability improvement from the Llama2 series to the Llama3 series. While our results on coarse length confirms the claimed length for all four tested models,\footnote{Since the double of coarse length exceeds the claimed length.} we observe a significant enhancement in fine length where the model perfectly remembers the copy target, notably between Llama-3-base and Llama-2-base (4k vs 0). 
Considering the similar model structures of Llama3 and Llama2, and the fact that Llama3's additional GQA~\cite{ainslie2023gqa} doesn't extend the context window, we attribute this improvement to Llama3's increased training data.

Our experiments validate the effectiveness of context-extending models based on transformer architectures~\cite{peng2024yarn,chen2023extending,bynamic_ntk,ntk_by_parts}. Their measured coarse lengths meet or exceed the claimed lengths, indicating superior long context modelling capabilities compared to original models.
Most of the long-sequence Transformers we examined were produced by increasing the theta in RoPE~\cite{su2024roformer} and then fine-tuning on long sequences; our empirical results suggest such strategies are generally effective approaches to extend transformer based LLM context. We argue that the main concern for transformer models to extend to long context remain the quadratic time and space complexity.\footnote{Our experiments on LWM (A model with 1M claimed context length and we show the forgetting curve partially at Appendix Figure~\ref{fig:LWM}) are very slow and suffers from out-of-memory (OOM) issues in our settings.}

Our measurement on RNN/SSM models show some negatives results. While RNNs theoretically support infinite context length, they exhibit a short coarse-grained Memory Length and zero fine-grained Memory Length, indicating an inability to perfectly memorize or to retain memory at any significant length. This highlights current limitations in RNN development for accurate token recall. Furthermore, in our measurement, RNNs like Mamba and RWKV exhibit higher language modeling accuracy than copy accuracy outside training context length (contrary to transformer models where two curves converge after certain length), suggesting an RNN model memory issue which seems to negatively affect correct token prediction at long context, see Appendix~\ref{appendix:forgetting_curves} for the Mamba and RWKV forgetting curves for more details.



\section{Analysis \& Discussion}
\label{sec:analysis}

\paragraph{Forgetting curve can be applied to a small model and is robust to data leakage.} We want to see whether the forgetting curve can be used to measure model memory for small models. To this end, we train an 83M parameter model using the Llama architecture on the PG-19 training dataset with a training context size of 1024. Further details are provided in Appendix~\ref{appendix:Llama-83m}. We then plot its forgetting curves, measured using both the PG-19 test and training datasets.

\begin{figure}
\centering
\includegraphics[width=0.48\textwidth]{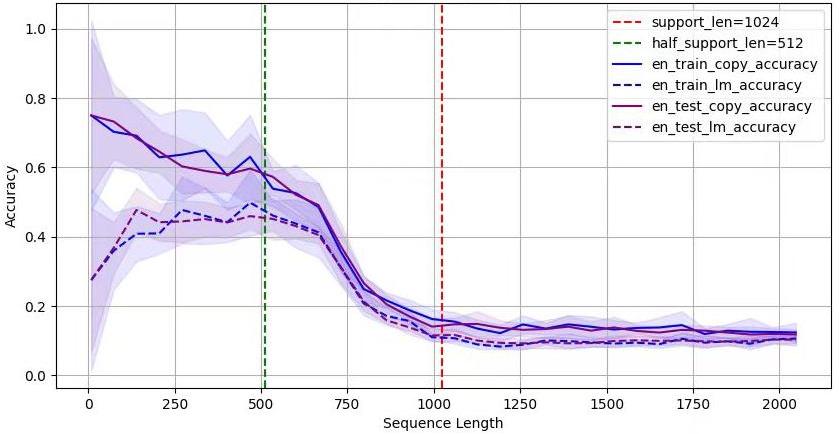}
\caption{The forgetting curves for the Llama model with 83M parameters. The solid and dashed lines represent copy accuracy and language modeling accuracy respectively. The model is trained on the PG-19 training dataset, and the forgetting curves are plotted using both the PG-19 training and test datasets.}
\label{fig:83m_forgetting_curve}
\end{figure}

Figure~\ref{fig:83m_forgetting_curve} shows the corresponding forgetting curves for the Llama-83M model. According to our fine memory and coarse memory definition presented in the previous section, the model does not have fine memory but clearly exhibit coarse memory slightly beyond half of the training context size. We remark that the measurement is achieved on such small and arguably under-trained models because our measurement could separate between the \textit{memory capability} and \textit{natural language understanding capability}. In the contrary, it is not possible to measure memory using popular metrics such as ``Needle in a Haystack'' as the model has not yet acquired understanding at this stage.  

We finally remark the closeness between forgetting curves measured on PG-19 test set and those measured on the training set. Obviously, both measurements point to very close coarse memory length. This shows that the forgetting curve is robust to data leakage as the forgetting curve greatly focuses on the difference between copy accuracy and LM accuracy.

\paragraph{Revisiting Perplexity using forgetting curve.}
Using the forgetting curve, we now revisit the hypothesis that perplexity is not directly related to better memorization or better usage of long context, which is first raised by~\cite{hu2024can}. 
We test this hypothesis using Transformer-XL~\cite{dai2019transformer}, which has shown perplexity improvement, often attributed to its superior utilization of long context.
We train a Transformer-XL style Llama (Llama-XL) on the PG-19 training set (details in Appendix ~\ref{appendix:Llama-XL}).

\begin{figure}[hbt!]
\centering
\includegraphics[width=0.48\textwidth]{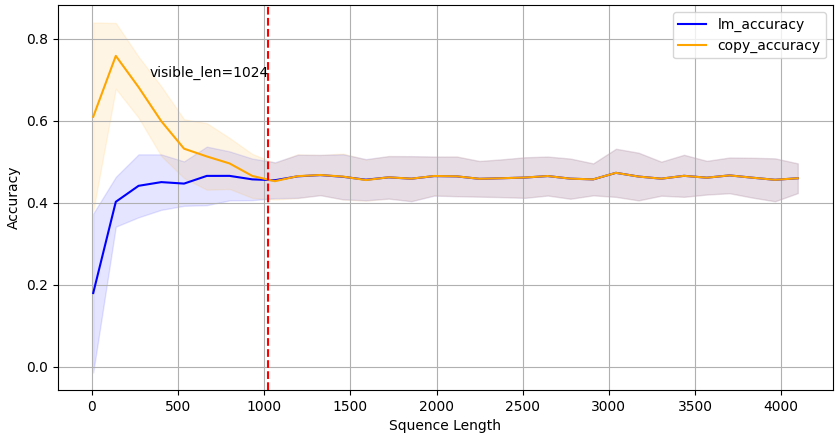}
\includegraphics[width=0.48\textwidth]{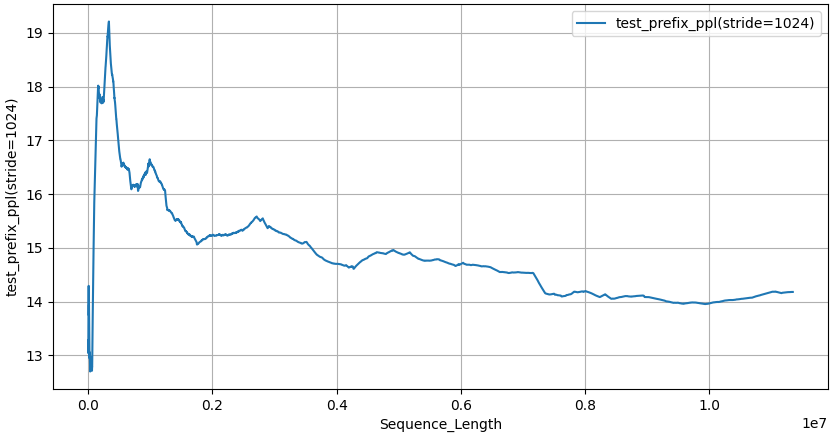}
\caption{The forgetting curve (top) and perplexity (bottom) for the Llama-XL, which is trained and tested on the PG-19 dataset.}
\label{fig:xl_forgetting_curve}
\end{figure}

We plot the forgetting curve as well as model perplexity curve in Figure~\ref{fig:xl_forgetting_curve} measured by PG-19 test set. For the perplexity curve, we confirm the findings of the original Transformer-XL paper where the architecture enables low perplexity for long context. However, the forgetting curve clearly shows that the perplexity performance is not related to model's memorization (or leverage) of long context, as copy curve becomes indistinguishable from the LM curve from 1k tokens while perplexity continues to decrease until several order of magnitude.\footnote{Anecdotally, Mistral v0.1 that uses sliding window attention (a technique that improves perplexity at long range) does not perform well on ``Needle in a Haystack'' at long range~\cite{online_1}.} Our empirical results support the hypothesis~\cite{hu2024can} that perplexity tests model's short context modelling capability and is not an appropriate measure to test model's long-context capability including memorization. 



\paragraph{Comparison of forgetting curve and other context length measurement.}

We show context length measured by forgetting curves as well as by Ruler~\cite{hsieh2024ruler} (i.e. the Effective Length column) in Table \ref{tab:main_results}. One can see that for models with low context length measured by forgetting curves (e.g. Mamba, RWKV), the effective length measured by Ruler is also low; however, the inverse does not hold, as shown by the experiments on ChatGLM3, Yarn-Llama-2 and LWM where we observe high context length while Ruler effective length remains low ($<$4k). We think this discrepancy are due to two interleaved factors: \ding{182} Ruler uses mainly synthetic tasks fulfilling which requires model's understanding capability, and \ding{183} to be counted as effective length in Ruler, the measured capability needs to exceed Llama-2-base/chat baseline. Contrary to Ruler's measurement, our measurement does not require understanding capability allowing us to measure model of any size (e.g. see Section~\ref{sec:analysis} for our measurement for a 83M model) and our method does not require a baseline model to define memory length. Fundamentally, we devise our approach trying to disentangle the understanding capability and the memory capability, and uniquely measure the later one.



\section{Conclusion}
In this work, we identify the limitations present in existing long-sequence language modeling metrics, tasks and benchmarks. To address these limitations, we introduce a new method called forgetting curve, which provides a flexible and robust measure to visualize LLM’s long-range memory capabilities. 

Using forgetting curve, we study the memory capabilities of fourteen open-source models with claimed context sizes ranging from 4K to 1M. 
Our findings empirically validate the effectiveness of existing transformer context extension methods, while raising questions for RNN/SSM architectures. Forgetting curve further reveals no direct correlation between memory capacity and perplexity. 

Compared to existing ways to measure long-context memory, forgetting curves mainly differs in trying to effectively decouple the model's long-term memory and language understanding ability. This provides a novel perspective for evaluating language models at all sizes, potentially guiding future research in model's long-term memory.


\section*{Limitations}
There are primarily two limitations in our work. Firstly, as the copy task requires two identical text segments, the maximum token dependency length is only half of the total sequence length, thus not achieving maximum efficiency for measurement. Notably, when the model imposes a strict limit on its maximum input length (e.g., ChatGLM3), our method allows only measuring up to half the maximum context length. Secondly, although we extensively tested the robustness under various sizes (83M, 7B), using different alignment settings (base and chat) and under various corpus settings, we remark that all the tests are performed within the Llama model family, leaving transferability of forgetting curves to other models unexplored.

\section*{Acknowledgments}
This work was supported in part by the National Science Foundation of China (No.62276056), the Natural Science Foundation of Liaoning Province of China (2022-KF-16-01), the Fundamental Research Funds for the Central Universities (Nos. N2216016 and N2316002), the Yunnan Fundamental Research Projects (No. 202401BC070021), and the Program of Introducing Talents of Discipline to Universities, Plan 111 (No.B16009).


\bibliography{custom}

\vfill
\appendix
\onecolumn
\newpage

\section{Statistical Analysis}
\label{appendix:statistical_analysis}
\begin{figure}[htp]
        \centering
        \includegraphics[width=0.8\textwidth]{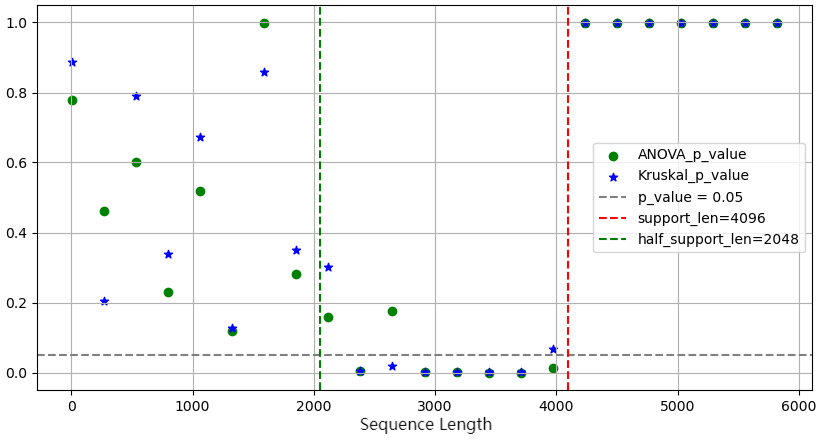} 
        \caption{ANOVA and Krustal  tests for LM accuracy.} 
        \label{fig:diff_irrelevant_t}
\end{figure}

In Figure \ref{fig:diff_irrelevant_t}, the one-way ANOVA tests and Kruskal-Wallis H-test were conducted with the copy target setting fixed at en\_test, and the irrelevant prefix set across three different settings: en\_test, en\_train, and zh. It can be seen that within half of the claimed length (2k), the choice of irrelevant prefix has no significant impact on the accuracy of language modeling.


\section{Models}
\label{appendix:models}
The models evaluated in section \ref{sec:examination} are listed in Table \ref{tab:models}.

\begin{table}[htp]
\small
    \centering
        \begin{tabular}{lccc}
        \hline
        \textbf{Name} & \textbf{Training
         length} & \textbf{Size} & \textbf{Huggingface~\cite{wolf2019huggingface} Source} \\
        \hline
        Llama-2-base~\cite{touvron2023llama} & 4k & 7B & meta-llama/Llama-2-7b-hf \\
        Llama-2-chat~\cite{touvron2023llama} & 4k & 7B & meta-llama/Llama-2-7b-chat-hf \\
        Llama-3-base~\cite{llama3_online} & 8k & 8B & meta-llama/Meta-Llama-3-8B \\
        Llama-3-chat~\cite{llama3_online} & 8k & 8B & meta-llama/Meta-Llama-3-8B-Instruct \\
        \hline
        Llama-2-base-32k~\cite{llama2_32k} & 32k & 7B & togethercomputer/LLaMA-2-7B-32K \\
        Mistral-base-v0.2~\cite{Mistral_online} & 32k & 7B & alpindale/Mistral-7B-v0.2-hf\\
        Mistral-chat-v0.2~\cite{Mistral_online} & 32k & 7B & mistralai/Mistral-7B-Instruct-v0.2 \\
        Mistral-chat-v0.3~\cite{Mistral_online} & 32k & 7B & mistralai/Mistral-7B-Instruct-v0.3 \\
        Qwen1.5~\cite{qwen1.5} & 32k & 7B & Qwen/Qwen1.5-7B \\
        ChatGLM3~\cite{zeng2022glm} & 128k & 6B & THUDM/chatglm3-6b-128k \\
        Yarn-Llama-2~\cite{peng2024yarn} & 128k & 7B & NousResearch/Yarn-Llama-2-7b-128k \\
        LWM~\cite{liu2023world} & 1M & 7B & LargeWorldModel/LWM-Text-1M \\
        \hline
        Mamba~\cite{gu2023mamba} & 2k & 2.8B & state-spaces/mamba-2.8b-hf \\
        RWKV~\cite{peng2023rwkv} & 4k & 7B & RWKV/v5-Eagle-7B-HF \\
        \hline
        \end{tabular}
    \caption{Information of evaluated and analyzed models.}
    \label{tab:models}
\end{table}

\section{Forgetting Curves of Open Source Models}
\label{appendix:forgetting_curves}
We have plotted the forgetting curves for all the open-source models listed in Table~\ref{tab:main_results}. 

\begin{figure*}[htp!]
    \centering
    \begin{minipage}{0.48\textwidth}
        \centering
        \includegraphics[width=\textwidth]{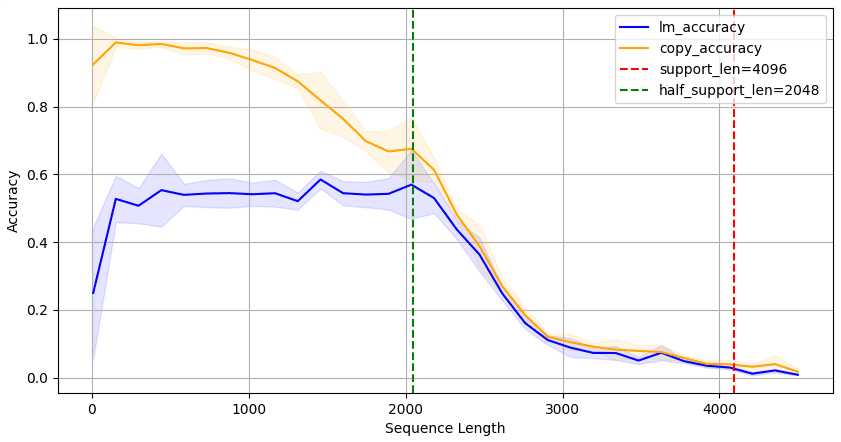}
        \caption{Llama-2-base}
        \label{fig:Llama-2-base}
    \end{minipage}%
    \hfill
    \begin{minipage}{0.48\textwidth}
        \centering
        \includegraphics[width=\textwidth]{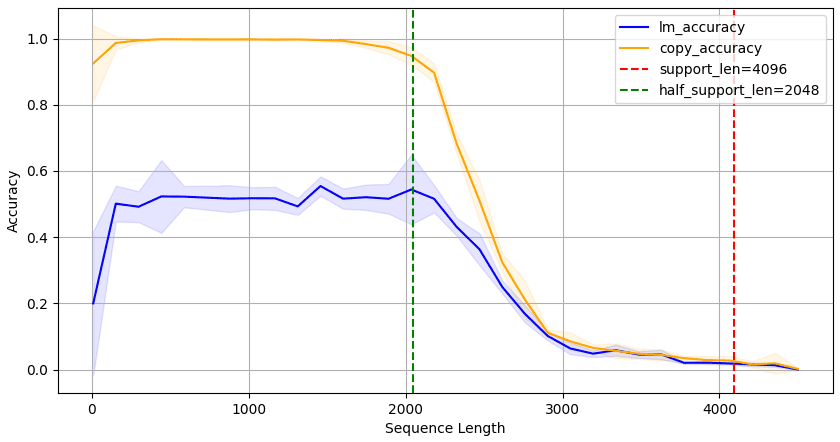}
        \caption{Llama-2-chat}
        \label{fig:Llama-2-chat}
    \end{minipage}
    \vfill
    \begin{minipage}{0.48\textwidth}
        \centering
        \includegraphics[width=\textwidth]{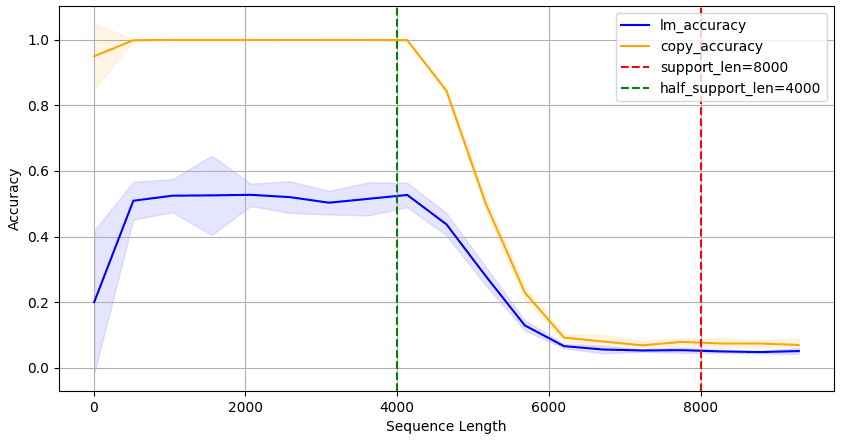}
        \caption{Llama-3-base}
        \label{fig:Llama-3-base}
    \end{minipage}%
    \hfill
    \begin{minipage}{0.48\textwidth}
        \centering
        \includegraphics[width=\textwidth]{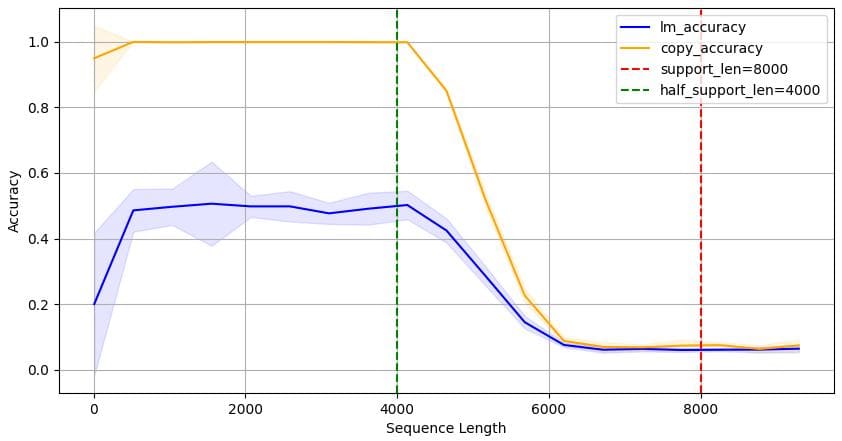}
        \caption{Llama-3-chat}
        \label{fig:Llama-3-chat}
    \end{minipage}
    \vfill
    \begin{minipage}{0.48\textwidth}
        \centering
        \includegraphics[width=\textwidth]{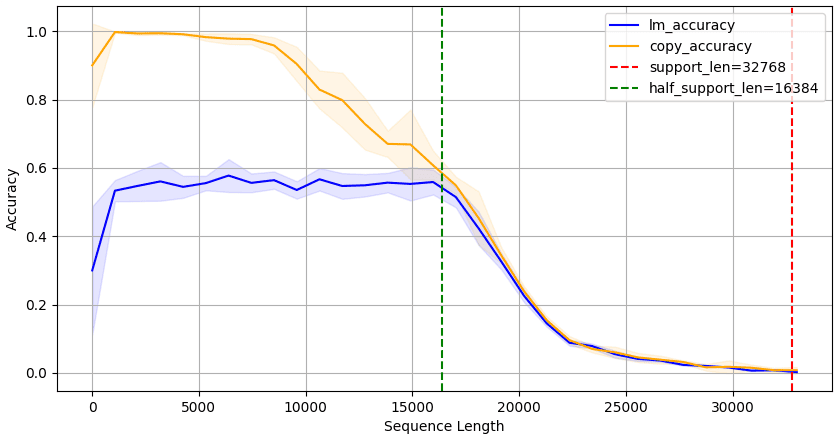}
        \caption{Llama-2-base-32k}
        \label{fig:Llama-2-base-32k}
    \end{minipage}%
    \hfill
    \begin{minipage}{0.48\textwidth}
        \centering
        \includegraphics[width=\textwidth]{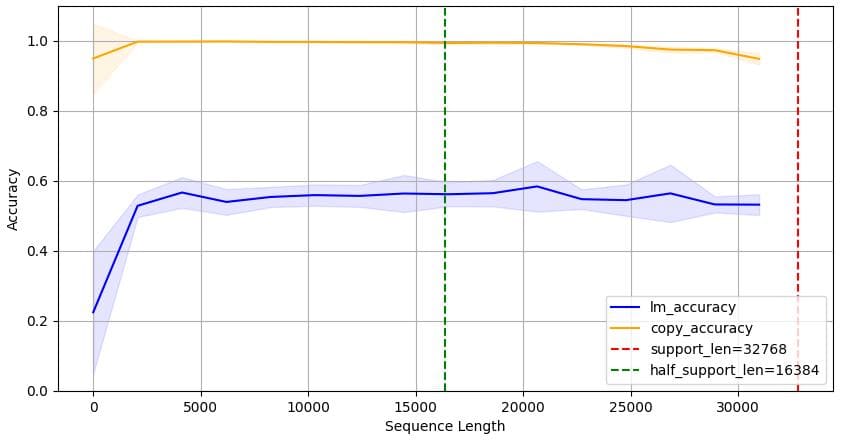}
        \caption{Mistral-base-v0.2}
        \label{fig:Mistral-base-v0.2}
    \end{minipage}
    \vfill
    \begin{minipage}{0.48\textwidth}
        \centering
        \includegraphics[width=\textwidth]{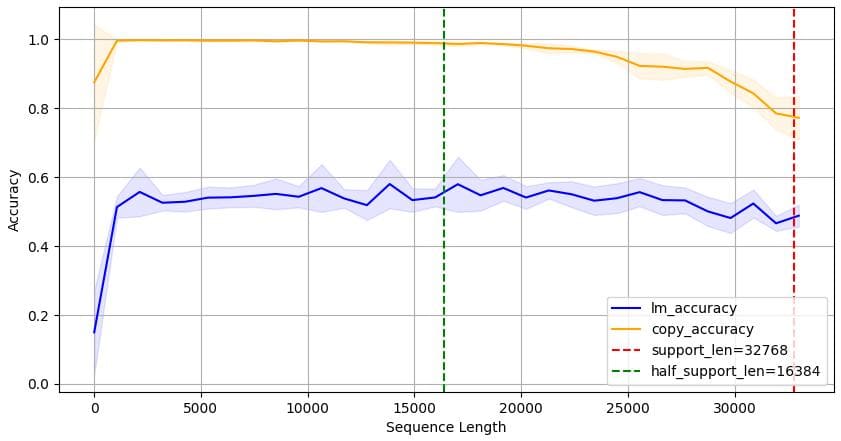}
        \caption{Mistral-chat-v0.2}
        \label{fig:Mistral-chat-v0.2}
    \end{minipage}%
    \hfill
    \begin{minipage}{0.48\textwidth}
        \centering
        \includegraphics[width=\textwidth]{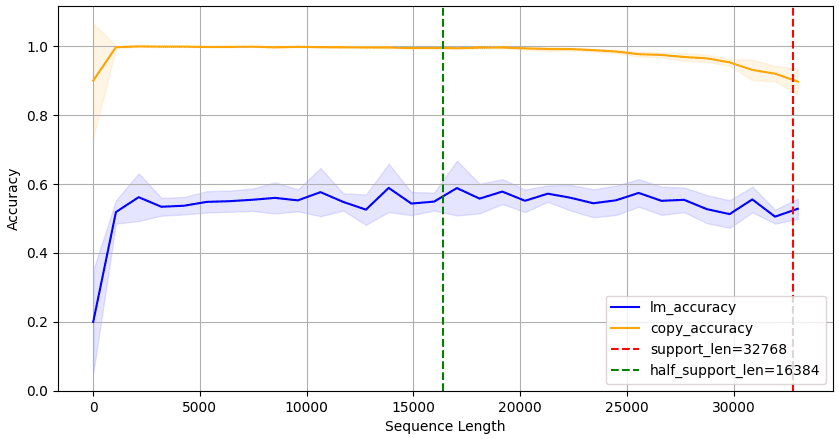}
        \caption{Mistral-chat-v0.3}
        \label{fig:Mistral-chat-v0.3}
    \end{minipage}
    \vfill
    \begin{minipage}{0.48\textwidth}
        \centering
        \includegraphics[width=\textwidth]{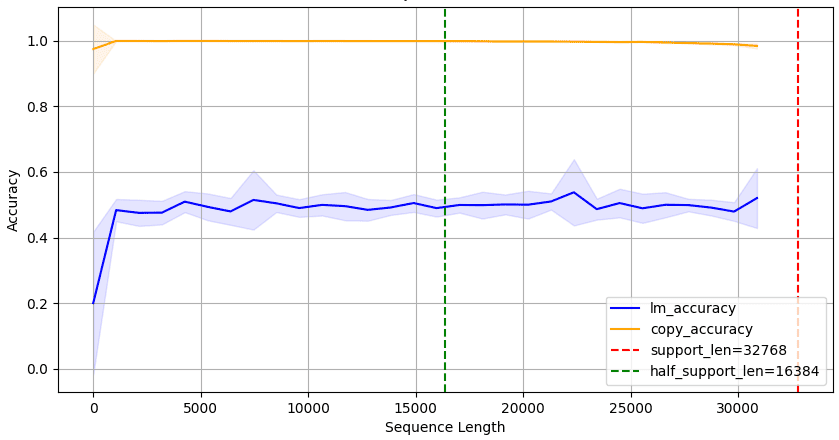}
        \caption{Qwen1.5}
        \label{fig:Qwen1.5}
    \end{minipage}%
    \hfill
    \begin{minipage}{0.48\textwidth}
        \centering
        \includegraphics[width=\textwidth]{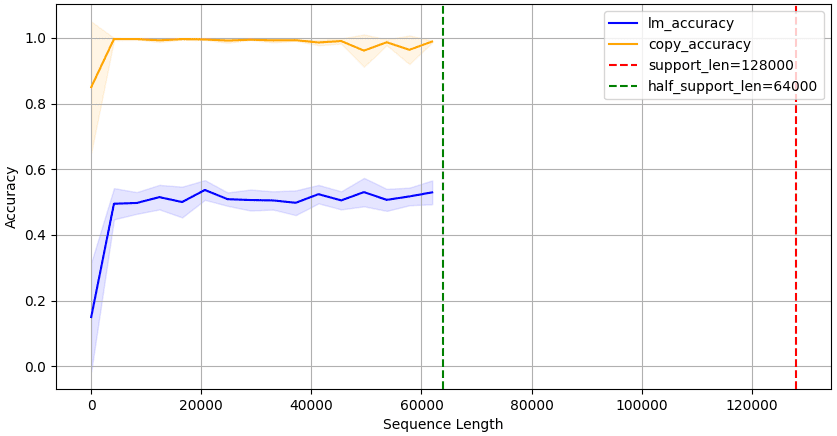}
        \caption{ChatGLM3}
        \label{fig:ChatGLM3}
    \end{minipage}
\end{figure*}

\begin{figure*}[htp!]
    \centering
    \begin{minipage}{0.48\textwidth}
        \centering
        \includegraphics[width=\textwidth]{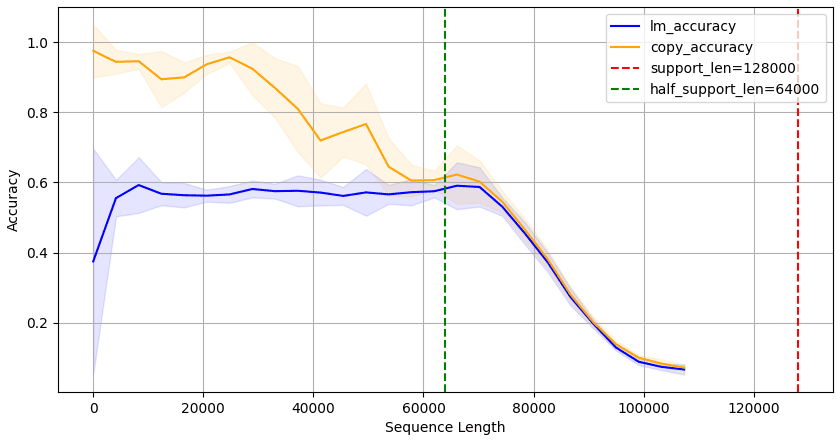}
        \caption{Yarn-Llama-2}
        \label{fig:Yarn-Llama-2}
    \end{minipage}%
    \hfill
    \begin{minipage}{0.48\textwidth}
        \centering
        \includegraphics[width=\textwidth]{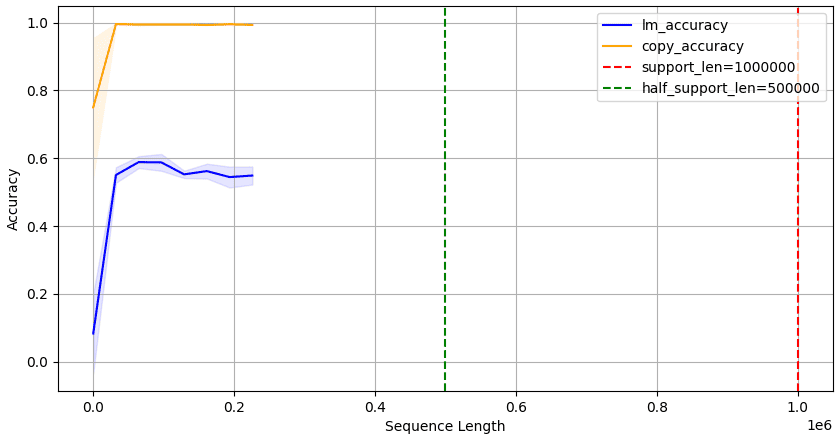}
        \caption{LWM}
        \label{fig:LWM}
    \end{minipage}%
    \vfill
    \begin{minipage}{0.48\textwidth}
        \centering
        \includegraphics[width=\textwidth]{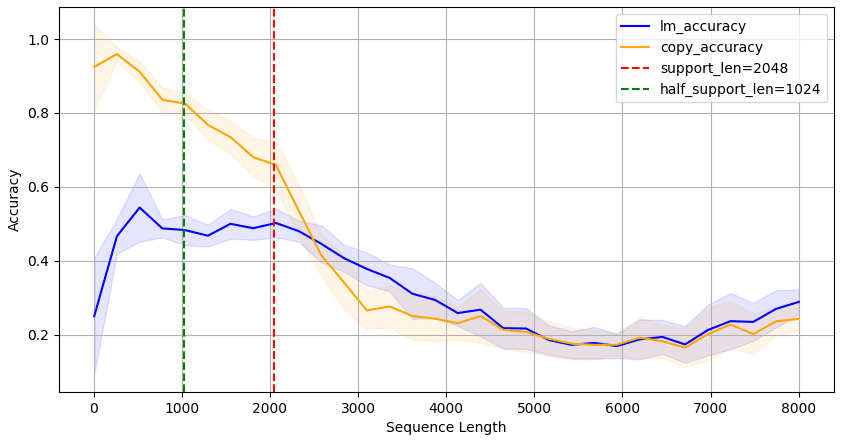}
        \caption{Mamba}
        \label{fig:Mamba}
    \end{minipage}
    \hfill
    \begin{minipage}{0.48\textwidth}
        \centering
        \includegraphics[width=\textwidth]{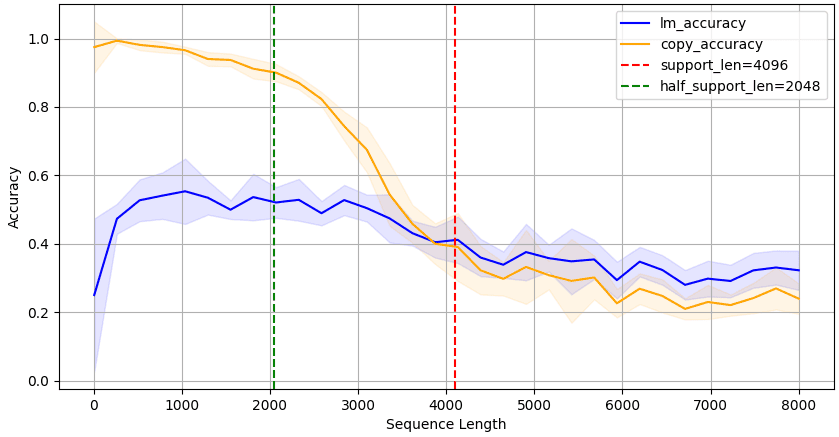}
        \caption{RWKV}
        \label{fig:RWKV}
    \end{minipage}
\end{figure*}

\newpage
\section{Details of Custom Model Training}
\label{appendix:model_detail}

\subsection{Llama-83M}
\label{appendix:Llama-83m}

\begin{figure}[htp]
    \centering
    \includegraphics[width=0.7 \linewidth]{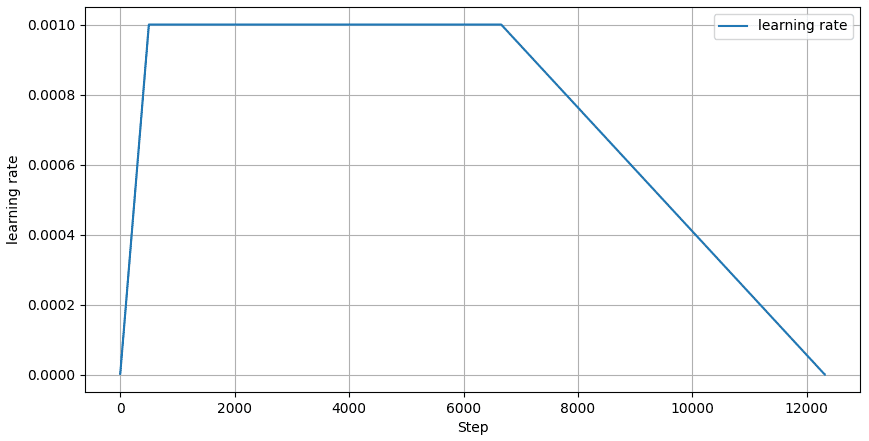}
    \caption{WSD scheduler}
    \label{fig:WSD_scheduler}
\end{figure}

We train the Llama-83M model for one epoch on the PG-19 training set, with the objective of maximizing the probability of the next token, which is the standard pre-training objective for language models. The model and training hyperparameters are listed in Table \ref{tab:hyperparameters}. We use the WSD scheduler~\cite{hu2024minicpm} shown in Figure \ref{fig:WSD_scheduler}. The optimization is performed using the AdamW optimizer~\cite{loshchilov2017decoupled} with parameters (beta1, beta2) = (0.9, 0.95), epsilon set to 1e-8, and a weight decay of 0.1. We do not employ dropout in our model.

\begin{table}[htp!]
\centering
    \centering
    \begin{tabular}{ll}
    \hline
    \textbf{Hyperparameters} & \textbf{Value} \\
    \hline
    model\_type & llama \\
    hidden\_act & silu \\
    initializer\_range & 0.02 \\
    hidden\_size & 512 \\
    intermediate\_size & 2048 \\
    max\_position\_embeddings & 2048 \\
    num\_attention\_heads & 8 \\
    num\_hidden\_layers & 12 \\
    num\_key\_value\_heads & 8 \\
    pretraining\_tp & 1 \\
    rms\_norm\_eps & $1.00 \times 10^{-5}$ \\
    tie\_word\_embeddings & FALSE \\
    torch\_dtype & float16 \\
    vocab\_size & 32000 \\
    \hline
    training\_len & 1024 \\
    total\_batch\_size & 256 \\
    batch\_size\_per\_device & 8 \\
    device\_count & 4 \\
    gradient\_accumulation\_steps & 8 \\
    learning\_rate & 0.001 \\
    max\_grad\_norm & 1 \\
    \hline
    \end{tabular}
    \caption{Model Hyperparameters and Training Hyperparameters.}
    \label{tab:hyperparameters}
\end{table}

\newpage
\subsection{Llama-XL}
\label{appendix:Llama-XL}

We modify the attention module of Llama to make it a model in the style of Transformer-XL~\cite{dai2019transformer}. Here is the attention formula for the $n$-th layer and the $\tau$-th segment:


\begin{eqnarray}
q^{n}_{\tau}, k^{n}_{\tau}, v^{n}_{\tau} & = & h^{n-1}_{\tau}W^{n}_{q}, h^{n-1}_{\tau}W^{n}_{k}, h^{n-1}_{\tau}W^{n}_{v} \\
\tilde{k}^{n}_{\tau} & = & L2Norm(k^{n}_{\tau}) \\
\hat{k}^{n}_{\tau}, \hat{v}^{n}_{\tau} & = & [k\_cache^n; \tilde{k}^{n}_{\tau}], [v\_cache^n; v^{n}_{\tau}]\label{eq:kv_cache_concatenation} \\
k\_cache^{n}, v\_cache^{n}& = & \tilde{k}^{n}_{\tau},v^{n}_{\tau} \\
pids & = & [0,1,2,...,len(\hat{k}^{n}_{\tau})-1] \\
\bar{q}^{n}_{\tau}, \bar{k}^{n}_{\tau} & = & apply\_rope(q^{n}_{\tau}, pids[-len(q^{n}_{\tau}):]), apply\_rope(\hat{k}^{n}_{\tau}, pids) \\
o^{n}_{\tau} & = & softmax(\bar{q}^{n}_{\tau}\bar{k}^{n^T}_{\tau} + Mask)\hat{v}^{n}_{\tau} \\
\tilde{h}^{n}_{\tau} & = & o^{n}_{\tau}W^{n}_{o}
\end{eqnarray}

\begin{figure}[t]
    \begin{minipage}{0.45\linewidth}
        \centering
        \includegraphics[width=\linewidth]{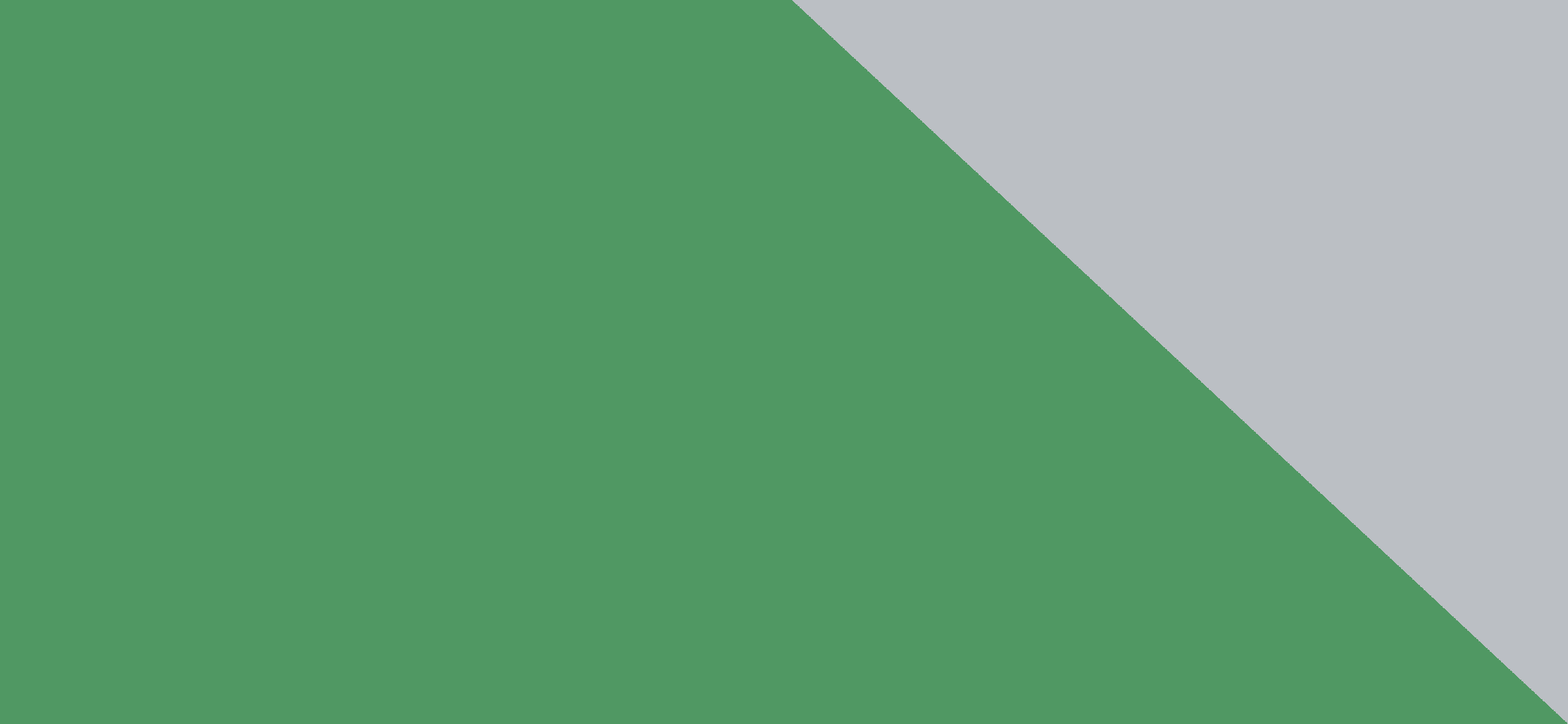}
        \caption{Mask used in Llama-XL. Elements in the green area are 0, and elements in the gray area are $-\infty$.}
        \label{fig:mask}
    \end{minipage}
    \hfill
    \begin{minipage}{0.50\linewidth}
        \centering
        \includegraphics[width=\linewidth]{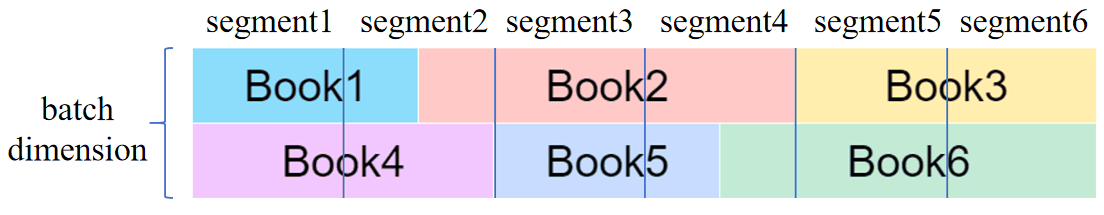}
        \caption{Ordered DataLoader.}
        \label{fig:data_loader}
    \end{minipage}
\end{figure}

$n$ denotes the layer index of the attention module.
$\tau$ denotes the segment index of the data.
$[;]$ denotes concatenation along the sequence dimension, while $[:]$ denotes slicing along the sequence dimension.
$h^{n-1}_{\tau}$ is the hidden state from the previous layer.
$W^{n}_{q}$, $W^{n}_{k}$, $W^{n}_{v}$, $W^{n}_{o}$ are the weight matrices for query, key, value, and output, respectively.
$q^{n}_{\tau}$, $k^{n}_{\tau}$, $v^{n}_{\tau}$ are the query, key, and value matrices.
$\tilde{k}^{n}_{\tau}$ is the key matrices after L2 normalization along the embedding dimension.
$\hat{k}^{n}_{\tau}$, $\hat{v}^{n}_{\tau}$ are the concatenated key and value matrices with the cached ones from the previous segment along the sequence dimension.
$k\_cache^{n}$, $v\_cache^{n}$ are the cached key and value matrices, updated with the current segment.
$pids$ is the sequence of position indices from 0 to $len(\hat{k}^{n}_{\tau})-1$.
$apply\_rope$ is the function that applies the rotary position embedding~\cite{su2024roformer}.
$Mask$ is the attention mask, which is a LowerTriangularFromBottomRightMask as illustrated in Figure~\ref{fig:mask}.
$o^{n}_{\tau}$ is the attention output.
$\tilde{h}^{n}_{\tau}$ is the output hidden state.

The main modification is in Equation (\ref{eq:kv_cache_concatenation}), where we concatenate the KV cache from the previous segment to the current one, unlike Transformer-XL, which concatenates the hidden state. It is worth noting that our position encoding spans from 0 to $len(\hat{k}^{n}_{\tau})-1$, thereby guaranteeing that the model employs position encodings encountered during training for inference. Normalizing the $k^{n}_{\tau}$ with L2Norm helps with the extrapolation of the position encoding~\cite{kexuefm-9859}, although we do not use position encoding outside of the training range. 
We also use an ordered DataLoader to ensure that segments are consecutive, as shown in Figure \ref{fig:data_loader}. Other experimental settings are consistent with Llama-83M.



\end{document}